\documentclass[conference]{IEEEtran}
\IEEEoverridecommandlockouts
\usepackage{cite}
\usepackage{amsmath,amssymb,mathabx}
\usepackage{amsfonts,mathtools,amsthm}
\usepackage{mathrsfs,bm,dsfont}
\usepackage{algorithmic}
\usepackage{graphicx}
\usepackage{multirow}
\usepackage{textcomp}
\usepackage{xcolor}
\usepackage{tikz}
\usepackage{comment}
\usetikzlibrary{arrows.meta}

\newtheorem{remark}{Remark}
\def\BibTeX{{\rm B\kern-.05em{\sc i\kern-.025em b}\kern-.08em
    T\kern-.1667em\lower.7ex\hbox{E}\kern-.125emX}}

\usepackage[
placement=top,
angle=0,
color=red,
scale=1.1,
hshift=0,
vshift=-15
]{background}
\backgroundsetup{contents=\bf{----------------------------------------------------------------- PREPRINT -----------------------------------------------------------------}}

\begin{document}

\title{Dynamic Control Allocation for\\ Dual-Tilt UAV Platforms\\ \thanks{*The results of this research were obtained in whole or in part as part of the Doctorate of National Interest in Robotics and Intelligent Machines. National PhD DRIM, Genova}}

\author{\IEEEauthorblockN{Marcello Sorge$^1$, Federico Ciresola$^1$, Giulia Michieletto$^{2,1}$, Angelo Cenedese$^{1,3}$}
\IEEEauthorblockA{\textit{$^1$Department of Information Engineering} \textit{University of Padova}
Padova, Italy\\
\textit{$^2$Department of Management and Engineering}, \textit{University of Padova}
Padova, Italy \\
\textit{$^3$Department of Industrial Engineering}, \textit{University of Padova}
Padova, Italy \\
corresponding contact: \texttt{marcello.sorge@unipd.it}}
}

\maketitle

\begin{abstract}
This paper focuses on dynamic control allocation for a hexarotor UAV platform, considering a trajectory tracking task as as case study. It is assumed that the platform is dual-tilting, meaning that it is able to tilt each propeller independently during flight, along two orthogonal axis. We present a hierarchical control structure composed of a high-level controller generating the required wrench for the tracking task, and a control allocation law ensuring that the actuators produce such wrench. The allocator imposes desired first-order dynamics on the actuators set, and exploits system redundancy to optimize the actuators state with respect to a given objective function. Unlike other studies on the subject, we explicitly model actuator saturation and provide theoretical insights on its effect on control performances. We also investigate the role of propeller tilt angles, by imposing asymmetric shapes in the objective function. Numerical simulations are presented to validate the allocation strategy.

\end{abstract}


\section{Introduction}
In the last decades, UAVs have been the subject of many researches, due to the extremely wide range of tasks they allow to perform. Among these, the most relevant include infrastructure inspection, environmental monitoring, aerial manipulation and search and rescue applications. Different from standard (fixed-propellers) multirotors, tilting  platforms are actuated by propellers whose spinning axes can change orientation during flight. This  enables omnidirectional vehicles capable of generating forces in arbitrary directions, independently of their body orientation. This feature is essential in interactive applications or challenging flights. 

The redundancy provided by the actuators' degrees of freedom complicates the control design because the mapping between the actuator inputs and the desired wrench is nonlinear and configuration-dependent. Furthermore, there are infinitely many control input trajectories that can produce the desired wrench on the platform. To address these challenges, control allocation strategies are often necessary. 
\medskip

\noindent{\textit{Related works:}} Several control allocation strategies for tilting multirotors have been developed. In \cite{static_diff_all}, a LQRI controller is proposed for jerk-level input allocation on a tilting multirotor. The actuator commands are obtained by multiplying the jerk-level commands by the Jacobian pseudo-inverse of the static allocation matrix. Projection onto the null space of such a matrix is exploited to maximize hover efficiency and handle kinematic and rank-reduction singularities. In \cite{multirotor}, a static differential allocation approach is proposed, in which actuator dynamics are assumed to be first-order linear. Projection onto the null space of the Jacobian of the static allocation matrix is combined with approximated propeller limit curves to keep the actuators within their saturation limits. \\
While many allocation strategies rely on static optimization, meaning that the optimal control input is computed at each control iteration, recently several dynamic control allocation strategies have been proposed, in which static optimization is replaced with gradient descent-like strategies. For a brief overview on dynamic control allocation, the  reader may refer to \cite{zaccarian_dynamic_allocation} and \cite{nonlinear_allocation}. In \cite{spline}, a dynamic control allocation strategy is proposed for systems subject to periodic reference trajectories. Optimality is considered with respect to a Lagrangian cost function, i.e., the objective function is defined as the integral of the steady-state periodic input signal over a period. A hybrid allocator is proposed, which updates the allocator state only after each period of the control input evolution is completed. This strategy is employed on a tilting quadrotor in \cite{spline_app}, where the wrench commanded by a high-level controller is converted into actuator velocities through a static allocator. The resulting control signal is fed to a dynamic allocator, based on a polynomial decomposition of the actuators transfer matrix, to minimize the spinning rate of a single, a priori chosen propeller. \\
However, many works on control allocation for tilting UAVs either neglect actuator dynamics or assume that they obey simple, linear, first-order dynamics. Instead, in \cite{rospo_theory}, a dynamic allocation approach is proposed, in which the actuator system is explicitly modeled as a first-order nonlinear redundant system. In \cite{rospo_theory}, a hierarchical structure is proposed, in which a high-level controller provides the desired wrench for the platform, while the allocator imposes some desired dynamics on the actuators state variables, ensuring that the output of such a system asymptotically converges to the control signal computed by the high-level controller. At the same time, the allocator injects in the actuators control input a component driving the actuators' state variables to the optimum with respect to a given objective function. Projection onto the null space of the Jacobian of the actuators output function ensures that optimization does not influence the behavior induced by the high-level controller. The allocation structure in \cite{rospo_theory} has been applied in \cite{rospo_2017} and in \cite{rospo_app} to the ROtor graSPing Omnidirectional (ROSPO) platform, an overactuated propeller-based ground platform.  In these works, the objective function aims at maintaining the actuators state within its saturation limits, while penalizing high propeller spinning rates in order to reduce energy consumption.
\medskip

\noindent{\textit{Contributions:}} In this work, we consider the allocator structure designed for the ROSPO platform and employ it to a double tilting UAV platform. Differently from previous works, we explicitly model saturation in the actuators output function, studying how it affects control performances. We also study the role of the propeller tilt angles, by defining objective functions that are asymmetric on tilt angles. The main contributions of this work can be summarized as follows:
\begin{itemize}
    \item we show how the allocator structure presented in \cite{rospo_theory}, \cite{rospo_2017} and \cite{rospo_app} can be adapted to a double-tilting UAV platform, explicitly considering actuator dynamics, and exploiting redundancy to optimize control performances with respect to a given secondary objective function;
    \item we provide theoretical insights on how actuator saturation affects control performance, and how the control structure leverages on the allocator to maintain the actuator state variables within their saturation limits at steady-state;
    \item we investigate the role of the propeller tilt angles in tracking tasks, by defining additional objective functions with asymmetric shapes in $\alpha$ and $\beta$, imposing different constraints on how propellers can tilt. 
\end{itemize}

The remainder of the paper is organized as follows: Section \ref{model} presents the dynamic equations describing the UAV platform and the actuators set. Section \ref{control} derives the high-level control law and describes the allocator equations from \cite{rospo_2017}-\cite{rospo_app}, highlighting convergence properties and the effect of saturation on the allocator performances. Section \ref{res} presents and discusses the results obtained through numerical simulations, and Section \ref{conclusion} summarizes the main results.
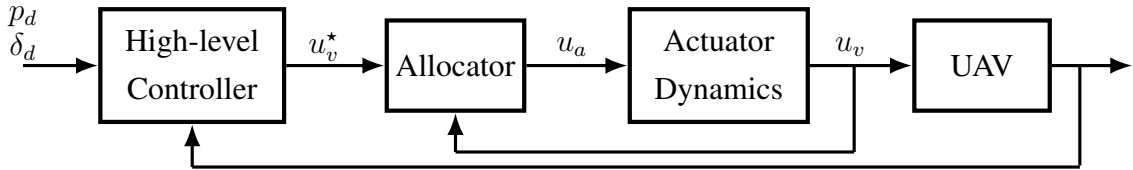
\begin{figure*}[t]
    \centering
    \begin{tikzpicture}
        \node (hlc) at (-7,0) [draw,rectangle,ultra thick,minimum height=1.2cm,minimum width=1.8cm]{\begin{tabular}{c}
            \large{High-level} \\
            \large{Controller}
        \end{tabular}};
        \node (all) at (-3.5,0) [draw,rectangle,ultra thick,minimum height=1.2cm,minimum width=1.8cm]{\large{Allocator}};
        \node (act) at (0,0) [draw,rectangle,ultra thick,minimum height=1.2cm,minimum width=1.8cm]{\begin{tabular}{c}
            \large{Actuator}  \\
            \large{Dynamics} 
        \end{tabular}};
        \node (uav) at (3.5,0) [draw,rectangle,ultra thick,minimum height=1.2cm,minimum width=1.8cm]{\large{UAV}};
        \draw[very thick,-{Latex}] (-9.25,0)--(hlc.west);
        \node at (-9.25,.25){\large{$\delta_d$}};
        \node at (-9.25,.65){\large{$p_d$}};
        \draw[very thick,-{Latex}] (hlc.east)--(all.west);
        \node at (-5.25,.25){\large{$u_v^{\star}$}};
        \draw[very thick,-{Latex}] (all.east)--(act.west);
        \node at (-1.95,.25){\large{$u_a$}};
        \draw[very thick,-{Latex}] (act.east)--(uav.west);
        \node at (1.75,.25){\large{$u_v$}};
        \draw[very thick] (1.8,0)--(1.8,-1.15);
        \draw[very thick] (1.8,-1.15)--(-3.5,-1.15);
        \draw[very thick,-{Latex}] (-3.5,-1.15)--(all.south);
        \draw[very thick,-{Latex}] (uav.east)--(5.5,0);
        \draw[very thick] (4.8,0)--(4.8,-1.33);
        \draw[very thick] (4.8,-1.35)--(-7,-1.35);
        \draw[very thick,-{Latex}] (-7,-1.35)--(hlc.south);
    \end{tikzpicture}
    \caption{Hierarchical control architecture showing the high-level controller and the low-level allocator}
    \label{fig:scheme}
\end{figure*}

\section{Dual-Tilt UAV Model} \label{model}

In this section, we derive the dynamics model for both a dual-tilt platform and the set of its actuators. Note that the UAV equations of motion describe a fully actuated system  whose six degrees of freedom (DoFs) are controlled by the \textit{input wrench} $u_v \in \mathbb{R}^6$, while redundancy is provided by the actuator set, which offers an infinite number of configurations compatible with a given reference wrench.

\subsection{Platform Dynamics}

Let $\mathscr{F}_W = \{O_{W},(x_{W},y_{W},z_{W})\}$ be the inertial reference frame (\textit{world frame)} and $\mathscr{F}_B = \{O_{B},(x_{B},y_{B},z_{B})\}$ be the reference frame whose origin is located at the UAV center of mass (CoM)(\textit{body frame}). Then, adopting the Euler-Newton approach, the dual-tilt platform dynamics is governed by the following equations of motion: 
\begin{equation}
    \begin{bmatrix}
        m\mathbf{\Ddot{p}}^{W}\\
        \mathbf{J}\boldsymbol{\Dot{\omega}^B}
    \end{bmatrix} = 
    \begin{bmatrix}
        \mathbf{R}_{WB}(\boldsymbol{\delta}) \mathbf{f}^B \\
        \boldsymbol{\tau}^B
    \end{bmatrix} -
    \begin{bmatrix}
        mg\hat{\mathbf{e}_3} \\
        \boldsymbol{\omega}^B \times \mathbf{J} \boldsymbol{\omega}^B
    \end{bmatrix}
    \label{uav_model_w}
\end{equation}
where $\mathbf{p}^W = \left[ x\;y\;z \right]^\top \in \mathbb{R}^3$ is the position of the CoM of the multirotor in $\mathscr{F}_W$, $\mathbf{R}_{WB}(\boldsymbol{\delta}) \in \mathbb{SO}(3)$ is the rotation matrix describing the relative orientation of $\mathscr{F}_B$ with respect to $\mathscr{F}_W$ and depending on the triplet of roll, pitch and yaw Euler angles  $\boldsymbol{\delta} = \left[\rho\; \theta\; \psi\right]^\top \in \mathbb{R}^3$, 
$m$ and $\mathbf{J} \in \mathbb{R}^{3 \times3}$ are respectively the multirotor mass and inertia tensor. Furthermore, $\mathbf{f}^B \in \mathbb{R}^3$ and $\boldsymbol{\tau}^B \in \mathbb{R}^3$ are the wrench components induced by the actuators, $\hat{\mathbf{e}_3} \in \mathbb{R}^3$ is the third canonical vector and $g = 9.81 m/s^2$ is the gravitational acceleration constant. Finally, $\boldsymbol{\omega}^B \in \mathbb{R}^3$ is the platform angular velocity expressed in $\mathscr{F}_B$ that can be expressed as a function of the Euler angles derivatives $\boldsymbol{\Dot{\delta}}$ as
\begin{equation}
    \boldsymbol{\omega}^B = \mathbf{W}(\boldsymbol{\delta})\boldsymbol{\Dot{\delta}} = 
    \begin{bmatrix}
        1 & 0 & -s\theta \\
        0 & c\rho & c\theta s\rho \\
        0 & -s\rho & c\theta c\rho
    \end{bmatrix} 
    \begin{bmatrix}
        \Dot{\rho} \\
        \Dot{\theta} \\
        \Dot{\psi}
    \end{bmatrix}
    \label{wtodeltadot}
\end{equation}
where $c$ and $s$ is a short notation to represent respectively $\cos{\cdot}$ and $\sin{\cdot}$, and matrix $\mathbf{W}(\boldsymbol{\delta}) \in \mathbb{R}^{3 \times 3}$ is invertible for $\theta \neq \pm \frac{\pi}{2}$.

\subsection{Actuators Dynamics}
Let $\mathscr{F}_i = \{O_i,(x_i,y_i,z_i)\}$ the frame whose origin is located in the $i$-th propeller spinning center and the $z_i$ axis is aligned with the propeller spinning axis, $i \in [1,6]$. The position $\mathbf{p}_i \in \mathbb{R}^3$ of the $i$-th propeller center in $\mathscr{F}_B$ and the orientation of its spinning axes $\mathbf{z}_i \in \mathbb{R}^3$ are respectively identified by 
\begin{align}
    \mathbf{p}_i &= \mathbf{R}_z(\gamma_i)\ell\hat{\mathbf{e}}_1
    \label{propeller_position}
\\
    \mathbf{z}_i &= \mathbf{R}_z(\gamma_i)\mathbf{R}_y(\beta_i)\mathbf{R}_x(\alpha_i)\hat{\mathbf{e}}_3
    \label{propeller_orientation}
\end{align}
where $\ell \in \mathbb{R}_{>0}$ is the length of the platform arms, $\hat{\mathbf{e}}_1 \in \mathbb{R}^3$ is the first canonical vector, $\mathbf{R}_z(\gamma_i)$, $\mathbf{R}_y(\beta_i)$, and $\mathbf{R}_x(\alpha_i) \in \mathbb{SO}(3)$ represent canonical rotations of angles $\gamma_i$, $\beta_i$ and $\alpha_i$ around the $z$, $y$ and $x$ axes of the body frame $\mathscr{F}_B$, respectively. More in detail, the cant angle  $\alpha_i \in [-\pi,\pi)$ describes the spinning axes rotations along the arm direction, whereas the \textit{dihedral} angle $\beta_i \in [0, \pi)$ describes rotations along the direction orthogonal to the arm. 

By rotating at spinning rate $\omega_i \in \mathbb{R}_{\geq0}$, any $i$-th propeller generates a force $\mathbf{f}_i^B\in \mathbb{R}^3$ and a torque $ \boldsymbol{\tau}_i^B \in \mathbb{R}^3$ at the origin of the body frame, expressed as
\begin{align}
    \mathbf{f}_i^B &= kc_f\omega_i \vert \omega_i \vert \mathbf{z}_i\\
    \boldsymbol{\tau}_i^B &= -\kappa c_{\tau}\omega_i \vert \omega_i \vert \mathbf{z}_i + \mathbf{p}_i \times \mathbf{f}_i^B
\end{align}
where $c_f$ and $c_{\tau}$ are the force and torque coefficients respectively, and $\kappa = 1$ for counter-clockwise rotating propellers, while $\kappa = -1$ for clockwise rotating propellers. Then, the total forces and torques exerted by the propellers on the platform CoM can be expressed as
\begin{equation}
    \begin{aligned}
        \mathbf{f}^B & = \sum_{i=1}^6 \mathbf{f}_i^B \\
        \boldsymbol{\tau}^B & = \sum_{i=1}^6 \boldsymbol{\tau}_i^B
    \end{aligned}
    \label{total_force_torque}
\end{equation}

In this work, the attention is focused on star-shaped dual-tilt hexarotors, thus we have that $\gamma_i$ in~\eqref{propeller_position} is equal to $(i-1)\pi/3$, $i \in [1,6]$ and $\alpha_i$ and $\beta_i$ in~\eqref{propeller_orientation} are time-varying parameters and constitute independently controllable inputs of the systems. Let $\mathbf{x}_a = \left[ \alpha_1,\cdots,\alpha_6,\beta_1,\cdots,\beta_6,\omega_1,\cdots,\omega_6 \right] ^\top \in \mathbb{R}^{18}$ be the \textit{state of the actuation system} whose components are subject to saturation limits, namely
\begin{equation}
    \underline{x}_{a,j} \leq x_{a,j} \leq \overline{x}_{a,j}, \quad j \in [1,18]
    \label{sat}
\end{equation}
We assume that it is impossible for the actuators' state to change instantaneously. Instead, the actuator set obeys first-order nonlinear dynamics $\Sigma_A$ described by
\begin{equation}
    \Sigma_A:
    \begin{cases}
        \dot{\mathbf{x}}_a = \mathbf{f}_a(\mathbf{x}_a) + \mathbf{g}_a(\mathbf{x}_a)\mathbf{u}_a \\
        \mathbf{y}_a= \mathbf{h}_a(\text{\textbf{sat}}(\mathbf{x}_a)) = \mathbf{u}_v
    \end{cases}
    \label{actuator_eq}
\end{equation}
where $\mathbf{u}_a \in \mathbb{R}^{18}$ is the \textit{actuators control signal} and  $\mathbf{u}_v = ((\mathbf{f}^B)^\top, (\boldsymbol{\tau}^B)^\top)^\top$ is the wrench acting on the platform CoM, defined in \eqref{total_force_torque}. Moreover, we have that $\mathbf{f}_a(\cdot): \mathbb{R}^{18} \rightarrow \mathbb{R}^{18}$, $\mathbf{g}_a(\cdot):\mathbb{R}^{18} \rightarrow \mathbb{R}^{18 \times18}$, $h_a(\cdot):\mathbb{R}^{18}\rightarrow\mathbb{R}^6$  are suitably defined continuous functions, with $g_a(x_a)$ being full-rank, and 
$\text{\textbf{sat}}(\cdot):\mathbb{R}^{18}\rightarrow \mathbb{R}^{18}$ implements the saturation in~\eqref{sat}. Since the dimension of the output vector $\boldsymbol{y_a}$ is smaller than the dimension of the input vector $\mathbf{u}_a$, the actuators set is a redundant dynamical system, meaning that there exist an infinite number of actuator state configurations compatible with a desired wrench to be exerted.

\section{Control} \label{control}
In this section, the high-level control law and the actuators control and allocation strategies are derived. The high-level controller specifies the wrench so that a given state-space trajectory for the multirotor CoM is tracked, while the allocator drives the actuators state variables to satisfy such request, optimizing the state $\mathbf{x}_a$ with respect to a given objective function.

\subsection{High-Level Controller}
Given a reference trajectory specified as $\boldsymbol{p_d^W}$, $\boldsymbol{\Dot{p}_d^W}$, $\boldsymbol{\Ddot{p}_d^W}$, $\boldsymbol{\delta_d}$, $\boldsymbol{\Dot{\delta}_d}$, $\boldsymbol{\Ddot{\delta}_d} \in \mathbb{R}^3$, we define the position tracking error and the orientation tracking error respectively as 
\begin{equation}
    \begin{aligned}
        \boldsymbol{e_p} & = \boldsymbol{p_d^W} - \boldsymbol{p^W} \\
        \boldsymbol{e_{\delta}} & = \boldsymbol{\delta_d} - \boldsymbol{\delta}
    \end{aligned}
\end{equation}
A control law ensuring asymptotic tracking of the reference trajectory is then given as $\boldsymbol{u_v^{\star}} = \left[ (\boldsymbol{f_c^B})^\top, (\boldsymbol{\tau_c^B})^\top \right]^\top$, where
\begin{equation}
    \begin{aligned}
        \boldsymbol{f_c^B} & = m \mathbf{R}_{WB}(\boldsymbol{\delta})^\top(\boldsymbol{\Ddot{p}_d} + \boldsymbol{K_D} \boldsymbol{\Dot{e}_p} + \boldsymbol{K_P e_p} + g\hat{\mathbf{e}_3}) \\
        \boldsymbol{\tau_c^B} & = \boldsymbol{JW}(\boldsymbol{\delta})(\boldsymbol{\Ddot{\delta}_d} + \boldsymbol{K_{D,\delta}\Dot{e}_{\delta}}+\boldsymbol{K_{P,\delta}e_{\delta})} + \boldsymbol{J\Dot{W}}(\boldsymbol{\delta})\boldsymbol{\Dot{\delta}} \\
        & + \boldsymbol{JW}(\boldsymbol{\delta})\boldsymbol{\Dot{\delta}} \times \boldsymbol{JW}(\boldsymbol{\delta})\boldsymbol{\Dot{\delta}}
    \end{aligned}
\end{equation}
where $\boldsymbol{K_D}$, $\boldsymbol{K_P}$, $\boldsymbol{K_{D,\delta}}$ and $\boldsymbol{K_{P,\delta}} \in \mathbb{R}^{3 \times 3}$ are diagonal and positive definite control gains. Then, by substituting \eqref{wtodeltadot} into the dynamic model \eqref{uav_model_w}, the tracking error dynamics can be written as
\begin{equation}
    \begin{aligned}
        \boldsymbol{\Ddot{e}_p} + \boldsymbol{K_D \Dot{e}_p} + \boldsymbol{K_P e_p} & = \boldsymbol{0} \\
        \boldsymbol{\Ddot{e}_{\delta}} + \boldsymbol{K_{D,\delta} \Dot{e}_{\delta}} + \boldsymbol{K_{P,\delta} e_{\delta}} & = \boldsymbol{0}
    \end{aligned}
\end{equation}
from which it is clear that $\left[\boldsymbol{e_p}^\top, \boldsymbol{e_{\delta}}^\top\right]^\top = \boldsymbol{0}$ is a globally asymptotically stable equilibrium.

\subsection{Allocator}
The goal of the allocator is to drive the wrench induced by the actuators to the wrench required by the high-level controller, while driving the actuators state variables to an optimum value with respect to a given objective function. To this end, the control law for the actuators set is chosen as
\begin{equation}
    \mathbf{u}_a = \boldsymbol{g}(\mathbf{x}_a)^{-1} \big( -\boldsymbol{f}(\mathbf{x}_a) + \boldsymbol{u_y} - \boldsymbol{u_j} \big)
    \label{ua}
\end{equation}
where 
\begin{equation}
    \begin{aligned}
        \boldsymbol{u_y} & = \gamma_p (\nabla \boldsymbol{h}(\text{\textbf{sat}}(\mathbf{x}_a) \nabla \text{\textbf{sat}}(\mathbf{x}_a))^{\dagger} \big( \boldsymbol{u_{v,c}} -\mathbf{u}_v \big) \\
        \boldsymbol{u_j} & = \gamma_j (\nabla \boldsymbol{h}(\text{\textbf{sat}}(\mathbf{x}_a)) \nabla \text{\textbf{sat}}(\mathbf{x}_a))_{\bot} \nabla \text{\textbf{sat}}(\mathbf{x}_a)^\top \nabla J(\text{\textbf{sat}}(\mathbf{x}_a))
        \label{uyuj}
    \end{aligned}
\end{equation}
with $\nabla(\cdot)$ representing the gradient of a function, the symbol $(\cdot)^{\dagger}$ represents the right-pseudoinverse of a matrix, the symbol $(\cdot)_{\bot}$ represents the operator projecting vectors onto the null space of a matrix, $\gamma_p$, $\gamma_j$ are positive scalars and $J(\mathbf{x}_a): \mathbb{R}^{18} \rightarrow \mathbb{R}$ is the objective function to be optimized. The diagonal operator $\nabla \text{\textbf{sat}}(\mathbf{x}_a) \in \mathbb{R}^{18 \times 18}$ is defined as
\begin{equation}
    \nabla sat(x_{a,h}) := 
    \begin{cases}
        1 & \text{if} \hspace{2mm} \underline{x}_{a,j} \leq x_{a,h} \leq \overline{x}_{a,j} \\
        0 & \text{otherwise}
    \end{cases}
\end{equation}
Substituting \eqref{ua} in \eqref{actuator_eq} and taking the derivative with respect to time of $\mathbf{u}_v$ yields
\begin{equation}
    \tilde{\Sigma}_A:
    \begin{cases}
        \mathbf{\Dot{x}}_a = \boldsymbol{u_y} - \boldsymbol{u_j} \\
        \boldsymbol{\Dot{u}_v} = \gamma_p (\boldsymbol{u_{v,c}} - \mathbf{u}_v)
    \end{cases}
    \label{act_simp}
\end{equation}
The term $\boldsymbol{u_y}$ ensures that the actuators output $\mathbf{u}_v$ converges to the high-level controller's output $\boldsymbol{u_v^{\star}}$, while the term $\boldsymbol{u_j}$ drives the actuators state to the optimum value without affecting the actuators output. For the purpose of this work, $\boldsymbol{f}(\mathbf{x}_a) = \boldsymbol{0}$, $\boldsymbol{g}(\mathbf{x}_a) = \boldsymbol{I_{18}}$, with $\boldsymbol{I_n}$ indicating the identity matrix in $\mathbb{R}^{n \times n}$. \\
The term $\boldsymbol{u_{v,c}}$ in \eqref{uyuj} is selected as
\begin{equation}
    \boldsymbol{u_{v,c}} = \boldsymbol{B}^{-1} \big( \boldsymbol{\Dot{u}_v^{\star}} - \boldsymbol{Au_v^{\star}} \big) + \boldsymbol{K \Tilde{u}_v}
    \label{u_vc}
\end{equation}
where $\boldsymbol{B} = \gamma_p\boldsymbol{I_6}$, $\boldsymbol{A} = -\gamma_p\boldsymbol{I_6}$, $\boldsymbol{\Tilde{u}_v} = \mathbf{u}_v - \boldsymbol{u_v^{\star}}$ and $\boldsymbol{K} \in \mathbb{R}^{6 \times 6}$ is a control gain to be chosen so that $\boldsymbol{A} - \boldsymbol{BK}$ is Hurwitz. Indeed, by substituting \eqref{u_vc} into \eqref{act_simp}, the resulting wrench error dynamics can be expressed as
\begin{equation}
    \boldsymbol{\Dot{\Tilde{u}}_v} = (\boldsymbol{A}-\boldsymbol{BK})\boldsymbol{\Tilde{u}_v}
\end{equation}
and clearly $\boldsymbol{\Tilde{u}_v} = \boldsymbol{0}$ is a globally asymptotically stable equilibrium. The hierarchical control structure is depicted in Figure \ref{fig:scheme}.
\begin{remark}
    Asymptotic convergence of $\mathbf{u}_v$ to $\boldsymbol{u_v^{\star}}$ is not guaranteed if the actuators state variables reach their saturation limits, due to the matrix $\nabla \text{\textbf{sat}}(\mathbf{x}_a)$ not being invertible. This can happen if the desired trajectory is too fast for the actuation capabilities of the robot, or if the high-level controller is too aggressively tuned. In the following, the operator $\nabla \text{\textbf{sat}}(\mathbf{x}_a)$ is computed as
    \begin{equation}
        \nabla sat(x_{a,h}) = 
        \begin{cases}
            1 & \text{if} \hspace{2mm} \underline{x}_{a,j} \leq x_{a,h} \leq \overline{x}_{a,j} \\
            \epsilon & \text{otherwise}
        \end{cases}
    \end{equation}
    with $\epsilon = 0.001$ to avoid numerical issues in simulations.
    \label{remark1}
\end{remark}
For our application, the objective function to be optimized is defined as
\begin{equation}
    J(\mathbf{x}_a) = \sum_{i=1}^6 \mu_{\alpha} \left( \frac{\Tilde{\alpha}_i}{\Delta \alpha_i} \right)^6 + \mu_{\beta} \left( \frac{\Tilde{\beta}_i}{\Delta \beta_i} \right)^6 + \mu_{\omega} \omega_{i}^2
    \label{cost}
\end{equation}
where $\mu_{\alpha}$, $\mu_{\beta}$ and $\mu_{\omega_a}$ are positive scalars, and
\begin{equation}
    \begin{aligned}
        \Tilde{x}_{a,h} & = x_{a,h} - x_{m_{a,h}} \\
        x_{m_{a,h}} & = \frac{\overline{x}_{a,j} + \underline{x}_{a,j}}{2} \\
        \Delta x_{a,h} & = \overline{x}_{a,j} - \underline{x}_{a,j}
    \end{aligned}
\end{equation}
The first two terms in \eqref{cost} aim at maximizing the distance between the propeller tilt angles and their saturation limits, while the last term aims at minimizing the propeller spinning rates, in order to minimize the energy consumption of the platform.
\begin{remark}
    The injection of $\boldsymbol{u_j}$ in the actuators control $\mathbf{u}_a$ provides a sub-optimal solution. Projection of $\gamma_j \nabla \text{\textbf{sat}}(\mathbf{x}_a)^\top \nabla J(\text{\textbf{sat}}(\mathbf{x}_a))$ implicitly encodes a soft constraint on the optimality of the control law, in order to ensure that the response induced by the controller is identical with $\gamma_j = 0$ (no optimization) and $\gamma_j \neq 0$ (optimized control law). As established in Remark \ref{remark1}, in case the state $\mathbf{x}_a$ reaches saturation, the structure of the projection operator in \eqref{uyuj} changes, resulting in different optimized and non-optimized responses. In the tests presented in this paper, however, this difference is negligible.
\end{remark}

\section{Simulations and Results} \label{res}
In this section we present the results obtained with numerical simulations on a UAV platform, exploiting the hierarchical control scheme depicted in Figure \ref{fig:scheme}. The platform considered for the tests is a coplanar, star-shaped hexarotor with mass $m = 2kg$ and inertia tensor $\mathbf{J} = diag(0.0217,0.0217,0.04) kgm^2$. Each propeller arm has length $l = 0.246m$, and all propeller arms are displaced by $\gamma = 60^{\circ}$, with the first propeller arm aligned with axis $x_B$. Odd-numbered propellers are counter-clockwise-spinning, while even-numbered propeller are clockwise-spinning. The force and torque coefficients are respectively $c_f = 8.59 \times 10^{-6}$ and $c_{\tau} = 1.37 \times 10^{-7}$ for all propellers. For simulation purposes, it is assumed that $\underline{\alpha} = \underline{\beta} -30^{\circ}$, $\overline{\alpha} = \overline{\beta} = 30^{\circ}$ , and finally $ \vert \underline{\omega} \vert = 100 rad/s$, $\vert \overline{\omega} \vert = 1000 rad/s$ for each propeller. \\
We define a circular trajectory to be followed, as:
\begin{equation}
    \begin{cases}
        x_d(t) = r \cos{(c_dt)} \\
        y_d(t) = r \sin{(c_dt)} \\
        z_d(t) = 0 \\
        \boldsymbol{\delta_d} = \boldsymbol{0}
    \end{cases}
\end{equation}
with $r = 2m$ and $c_d = 0.8s^{-1}$. \\
For all simulations, the control parameters are chosen as $\boldsymbol{K_P} = \boldsymbol{K_{P,\alpha}} = 2\boldsymbol{I_3}$, $\boldsymbol{K_D = K_{D,\alpha}} = 1.5\boldsymbol{I_3}$, $\gamma_p = 5$ and $\boldsymbol{K} = 3\boldsymbol{I_6}$. The objective function parameters are set as $\mu_{\alpha} = \mu_{\beta} = 750$, $\mu_{\omega} = \frac{1}{200}$ for all propellers.\\
The platform at time $t = 0s$ is statically hovering at $\boldsymbol{p_0}^W = (2,0,0) $, with initial orientation $\boldsymbol{\delta_0} = \boldsymbol{0}rad$ and with all propeller in a collinear configuration ($\alpha_0 = \beta_0 = 0rad$ for all propellers), and $\omega_i = \pm \sqrt{\frac{mg}{6c_f}}$, depending on the spinning direction of propeller $i$.

\subsection{Non-optimized vs Optimized Response} \label{gamj}
First, we compare the performance of the control architecture with respect to the cost function defined in \eqref{cost}, setting respectively $\gamma_j = 0$ and $\gamma_j = 10$. 
\begin{figure*}[t]
    \centering
    \begin{minipage}{.32\textwidth}
        \includegraphics[width=\textwidth,height=.75\textwidth]{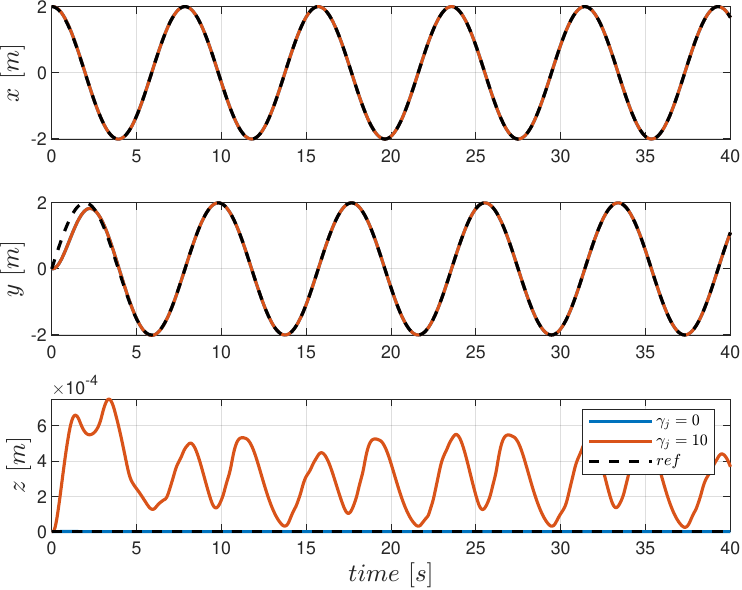}
        \caption{UAV position tracking}
        \label{fig:circ_pos}
    \end{minipage}
    \begin{minipage}{.32\textwidth}
        \includegraphics[width=\textwidth,height=.75\textwidth]{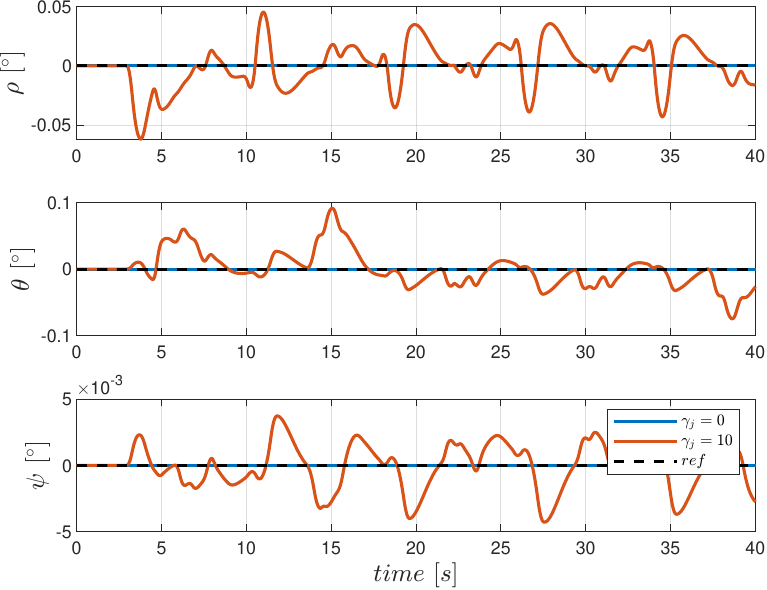}
        \caption{UAV Euler angles tracking}
        \label{fig:circ_angs}
    \end{minipage}
    \begin{minipage}{.32\textwidth}
        \includegraphics[width=\textwidth,height=.75\textwidth]{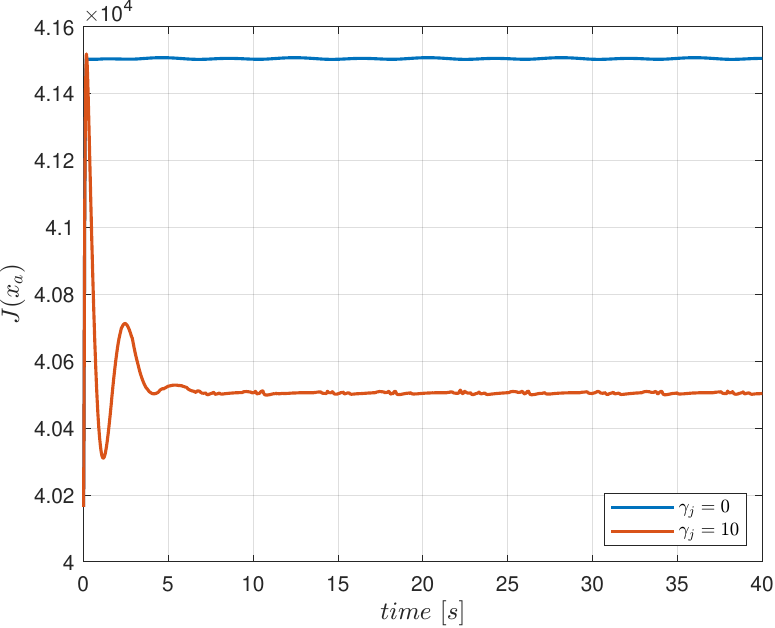}
        \caption{Objective functions}
        \label{fig:costs}
    \end{minipage}
\end{figure*}
\begin{figure*}[t]
    \centering
    \begin{minipage}{.49\textwidth}
    \includegraphics[width=.98\textwidth,height=.75\textwidth]{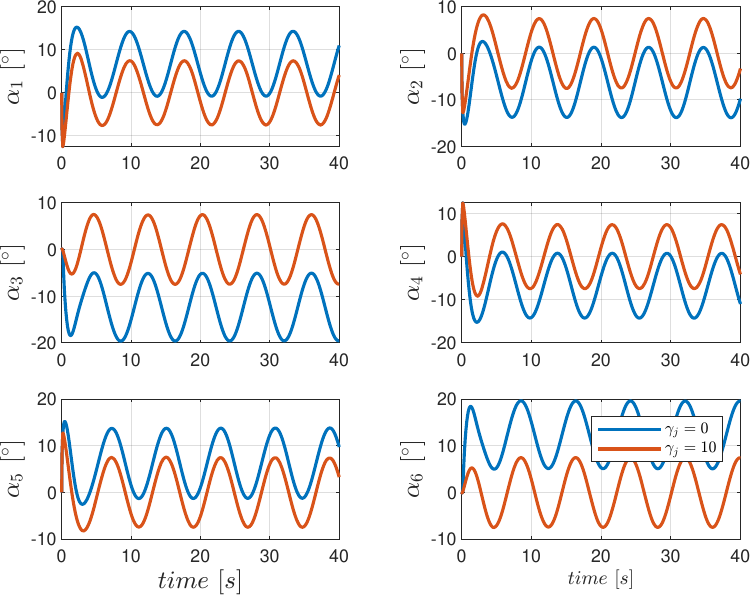}
        \caption{Propeller $\alpha$ angles}
        \label{fig:circ_alpha}
    \end{minipage}
    \begin{minipage}{.49\textwidth}
    \includegraphics[width=.98\textwidth,height=.75\textwidth]{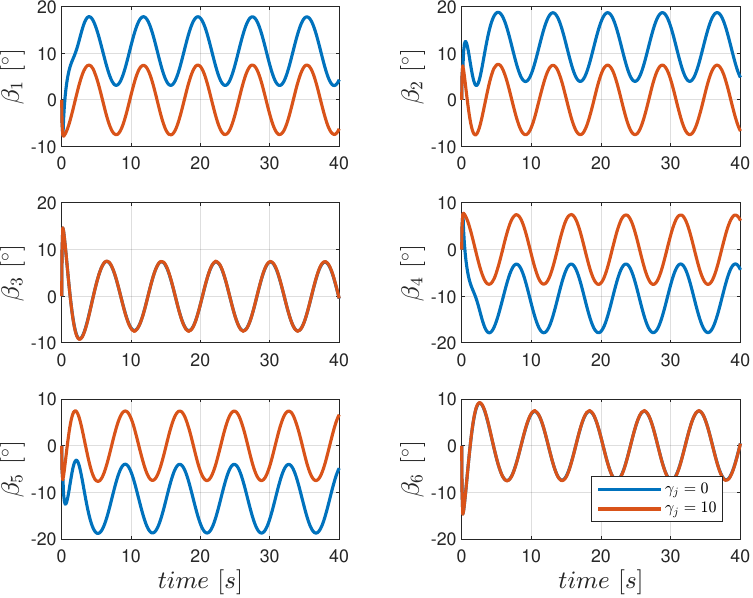}
        \caption{Propeller $\beta$ angles}
        \label{fig:circ_beta}
    \end{minipage}
\end{figure*}
\begin{figure}[t]
    \centering
    \begin{minipage}{.49\textwidth}
    \includegraphics[width=.98\textwidth,height=.75\textwidth]{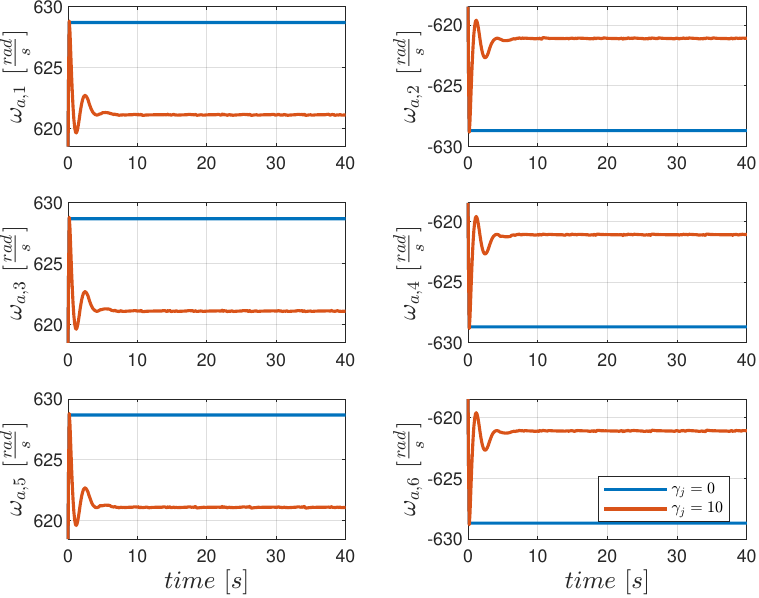}
        \caption{Propeller spinning rates}
        \label{fig:circ_wa}
    \end{minipage}
\end{figure}
The tracking performances can be observed in Figures \ref{fig:circ_pos} and \ref{fig:circ_angs}. Figures \ref{fig:circ_alpha}-\ref{fig:circ_beta} instead show the tilt angles for each propeller. It can be seen that introducing optimization in the allocator equations leads the propeller tilt angles to oscillate around zero (the middle of their saturation interval, according to the assumptions previously stated).
\begin{table}[htbp]
    \caption{Propeller tilt angles amplitudes and offsets}
    \centering
    \begin{tabular}{|c|c|c|}
        \hline
        \multirow{3}{4em}{\textbf{Propellers}} & \textbf{Amplitude} $[^{\circ}]$ & \textbf{Amplitude} $[^{\circ}]$\\
        & \textbf{Offset} $[^{\circ}]$ & \textbf{Offset} $[^{\circ}]$ \\
        & $\gamma_j = 0$ & $\gamma_j = 10$ \\
        \hline
        \multirow{2}{4em}{$[\alpha_1,\alpha_4]$} & $[7.5115,-7.5115]$ & $[7.2708,-7.2708]$ \\
        & $[6.7093,-6.7093]$ & $[0,0]$ \\
        \hline
        \multirow{2}{4em}{$[\alpha_2,\alpha_5]$} & $[7.5229,-7.5229]$ & $[7.413,-7.413]$ \\
        & $[-6.2338,6.2338]$ & $[0,0]$ \\
        \hline
        \multirow{2}{4em}{$[\alpha_3,\alpha_6]$} & $[7.2422,-7.2422]$ & $[7.437,-7.437]$ \\
        & $[-12.3415,12.3415]$ & $[0,0]$ \\
        \hline
        \multirow{2}{4em}{$[\beta_1,\beta_4]$} & $[7.315,-7.315]$ & $[7.4485,-7.4485]$ \\
        & $[10.5367,-10.5367]$ & $[0,0]$ \\
        \hline
        \multirow{2}{4em}{$[\beta_2,\beta_5]$} & $[-7.3453,7.3453]$ & $[-7.4485,7.4485]$ \\
        & $[11.4284,-11.4284]$ & $[0,0]$ \\
        \hline
        \multirow{2}{4em}{$[\beta_3,\beta_6]$} & $[-7.3568,7.3568]$ & $[-7.4427,7.4427]$ \\
        & $[-0.0245,0.0245]$ & $[0,0]$ \\
        \hline
    \end{tabular}
    \label{tab:amp_off_gamj}
\end{table}
To further analyze the platform behavior, we estimate the amplitude and offset (with respect to the middle of the saturation interval) of the tilt angles for each propeller at steady state (after $10s$), by fitting the data with a cosine function. Table \ref{tab:amp_off_gamj} shows such amplitudes and offsets for opposite propellers. It can be noticed how opposite propellers show an anti-symmetric behavior. Moreover, it is evident that the tilt angles oscillate around the middle of the saturation intervals. From figure \ref{fig:circ_wa}, it can be notice how the magnitude of the propeller spinning rates reduces when the term $u_j$ is active, namely $\gamma_j \neq 0$. Consequently, significant performance improvement with respect to the objective function $J(\mathbf{x}_a)$ can be observed in Figure \ref{fig:costs}. We highlight the fact that the objective function for the case $\gamma_j = 10$ shows a non-periodic behavior, despite the reference trajectory being periodic. This can be attributed to the fact that the objective function in \eqref{cost} is not integrated over a period of the trajectory. Finally, we notice that there is a small difference between the non-optimized and optimized UAV trajectories. This difference is however negligible for tracking purposes.
\begin{figure*}[t]
    \centering
    \begin{minipage}{.49\textwidth}
    \includegraphics[width=.98\textwidth,height=.75\textwidth]{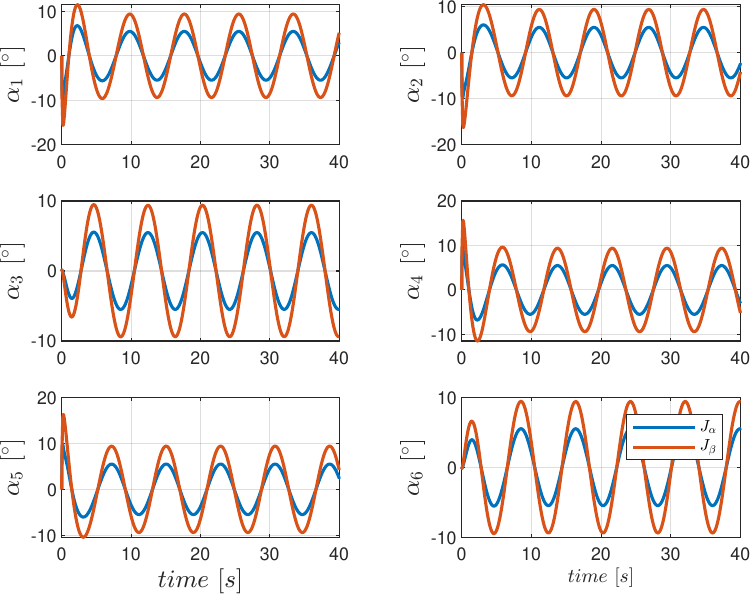}
        \caption{Propeller $\alpha$ angles}
        \label{fig:alpha_ab}
    \end{minipage}
    \begin{minipage}{.49\textwidth}
    \includegraphics[width=.98\textwidth,height=.75\textwidth]{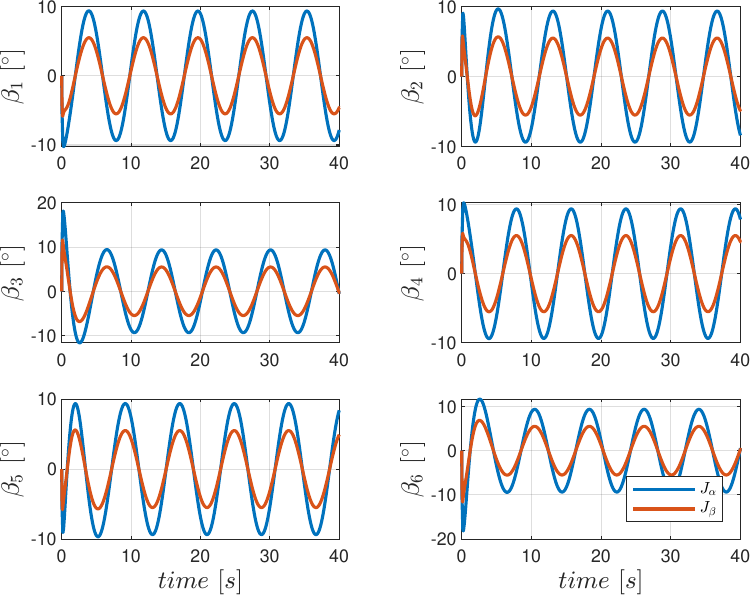}
        \caption{Propeller $\beta$ angles}
        \label{fig:beta_ab}
    \end{minipage}
\end{figure*}
\begin{figure*}[t]
    \centering
    \begin{minipage}{.49\textwidth}
    \includegraphics[width=.98\textwidth,height=.75\textwidth]{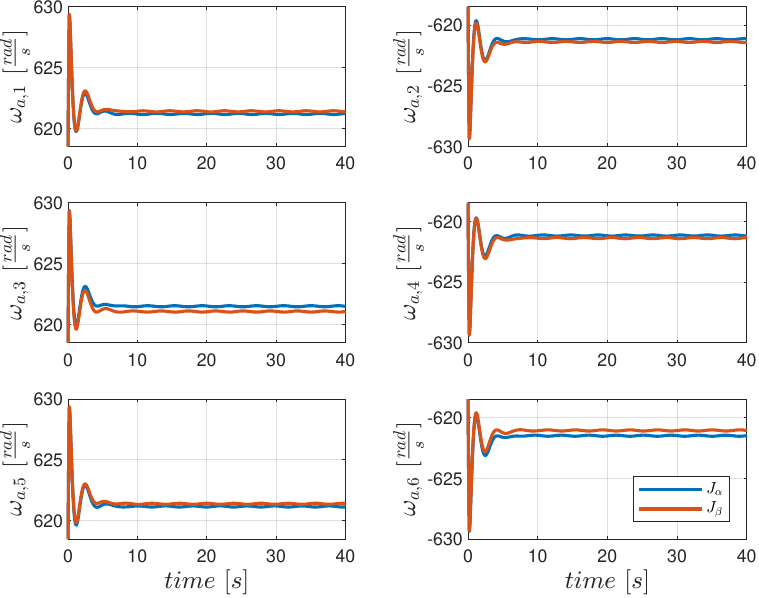}
        \caption{Propeller spinnin rates}
        \label{fig:omega_ab}
    \end{minipage}
    \begin{minipage}{.49\textwidth}
    \includegraphics[width=.98\textwidth,height=.75\textwidth]{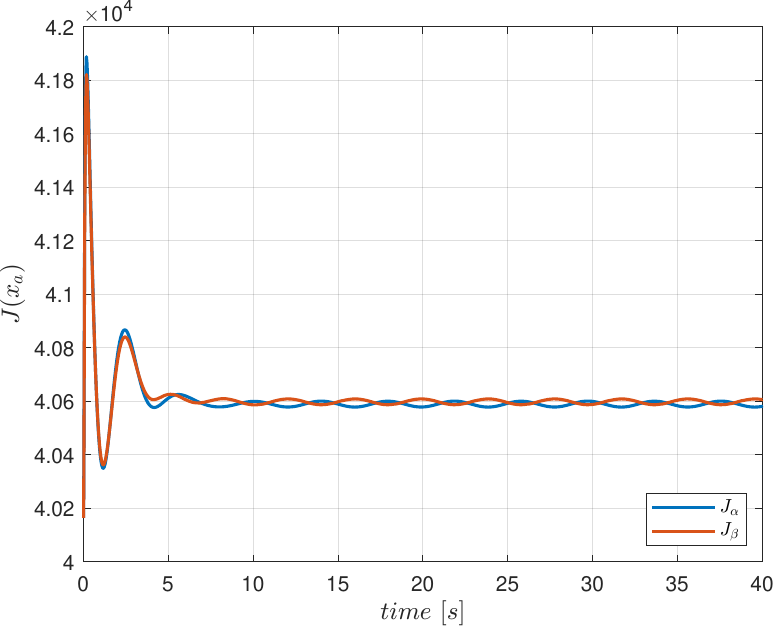}
        \caption{Propeller $\beta$ angles}
        \label{fig:cost_ab}
    \end{minipage}
\end{figure*}

\subsection{Asymmetric optimization on $\alpha$, $\beta$}
Next, we investigate the role of the tilt angles $\alpha$ and $\beta$, by defining two different objective functions with asymmetric shapes along $\alpha$ and $\beta$, namely:
\begin{equation}
    \begin{aligned}
        J_{\alpha}(\mathbf{x}_a) & = \sum_{i = 1}^6 \mu_{\alpha} \left( \frac{\Tilde{\alpha}_i}{\Delta \alpha_i} \right)^2 + \mu_{\beta} \left( \frac{\Tilde{\beta}_i}{\Delta \beta} \right)^6 + \mu_{\omega} \omega_{i}^2 \\
        J_{\beta}(\mathbf{x}_a) & = \sum_{i=1}^6 \mu_{\alpha} \left( \frac{\Tilde{\alpha}_i}{\Delta \alpha_i} \right)^6 + \mu_{\beta} \left( \frac{\Tilde{\beta}_i}{\Delta \beta_i} \right)^2 + \mu_{\omega} \omega_{i}^2
    \end{aligned}
\end{equation}
The objective function $J_{\alpha}(\mathbf{x}_a)$ has a steeper gradient along the $\alpha$ tilt angle compared to $J_{\beta}(\mathbf{x}_a)$. Consequently, the signal $\boldsymbol{u_j}$ will keep the $\alpha$ angle closer to the center of the saturation interval, allowing greater excursions for the $\beta$ tilt angles. The opposite holds considering $J_{\beta}(\mathbf{x}_a)$.
Figures \ref{fig:alpha_ab}, \ref{fig:beta_ab} and \ref{fig:omega_ab} show respectively the $\alpha$ and $\beta$ tilt angles and the spinning rates of each propellers. We repeat the amplitude estimation process in \ref{gamj}. For this analysis, we only estimate amplitudes since the offset with respect to zero is negligible.
\begin{table}[htbp]
    \caption{Propeller tilt angles amplitudes}
    \centering
    \begin{tabular}{|c|c|c|}
    \hline
        \multirow{2}{4em}{\textbf{Propellers}} & \textbf{Amplitude} $[^{\circ}]$ & \textbf{Amplitude} $[^{\circ}]$ \\
         & $J_{\alpha}$ & $J_{\beta}$ \\
         \hline
         $[\alpha_1,\alpha_4]$ & $[5.4786,-5.4786]$ & $[9.3793,-9.3793]$ \\
         \hline
         $[\alpha_2,\alpha_5]$ & $[5.4786,-5.4786]$ & $[9.385,-9.385]$ \\
         \hline
         $[\alpha_3,\alpha_6]$ & $[5.4826,-5.4826]$ & $[9.385,-9.385]$ \\
         \hline
         $[\beta_1,\beta_4]$ & $[9.385,-9.385]$ & $[5.4987,-5.4987]$ \\
         \hline
         $[\beta_2,\beta_5]$ & $[-9.385,9.385]$ & $[-5.4981,5.4987]$ \\
         \hline
         $[\beta_3,\beta_6]$ & $[-9.385,9.385]$ & $[-5.4489,5.4941]$ \\
         \hline
    \end{tabular}
    \label{tab:amp_jajb}
\end{table}
In table \ref{tab:amp_jajb} a significant difference in the amplitude of the propeller tilt angles can be observed comparing the results with $J(\mathbf{x}_a) = J_{\alpha}(\mathbf{x}_a)$ and $J(\mathbf{x}_a) = J_{\beta}(\mathbf{x}_a)$. We now study the effect of asymmetric optimization of the $\alpha$ and $\beta$ angles on the propeller spinning rates. Similarly to the previous analyses, the amplitude and offset obtained by fitting the spinning rates of each propeller at steady-state with a cosine wave are reported in table \ref{tab:omega_a}. It can be noticed that asymmetric optimization on $\alpha$ and $\beta$ has a minor impact on the propeller spinning rates. This shows that the tilt angles play an equivalent role during flight. Then, if there is any design or control constraint on either $\alpha$ or $\beta$, the allocator is able to preserve asymptotic tracking by acting on $\beta$ or $\alpha$, respectively, without significantly impacting the propeller spinning rates, and therefore the energy consumption of the platform.
\begin{table}[htbp]
    \caption{Propeller spinning rates amplitudes and offsets}
    \centering
    \begin{tabular}{|c|c|c|}
        \hline
        \multirow{3}{4em}{\textbf{Propellers}} & \textbf{Amplitude} $[\frac{rad}{s}]$ & \textbf{Amplitude} $[\frac{rad}{s}]$\\
        & \textbf{Offset} $[\frac{rad}{s}]$ & \textbf{Offset} $[\frac{rad}{s}]$ \\
        & $J_{\alpha}$ & $J_{\beta}$ \\
        \hline
        \multirow{2}{4em}{$[\omega_1,\omega_4]$} & $[0.0451,0.0451]$ & $[0.0455,0.0455]$ \\
        & $[621.1694,-621.1694]$ & $[621.3834,-621.3834]$ \\
        \hline
        \multirow{2}{4em}{$[\omega_2,\omega_5]$} & $[-0.0454,0.0454]$ & $[0.0434,-0.0435]$ \\
        & $[-621.1664,621.1664]$ & $[-621.387,621.3872]$ \\
        \hline
        \multirow{2}{4em}{$[\omega_3,\omega_6]$} & $[-0.0436,0.0436]$ & $[0.046,-0.0458]$ \\
        & $[621.4924,-621.4924]$ & $[621.0602,-621.0602]$ \\
        \hline
    \end{tabular}
    \label{tab:omega_a}
\end{table}

\section{Conclusions} \label{conclusion}
In this work, we have shown how the allocation architecture presented in \cite{rospo_2017} can be adapted to an aerial platform, and we demonstrated how such architecture can compensate eventual constraints on the propeller tilt angles with minor effect on the platform energy consumption. We also provided insights on how actuator saturation affects the control performance. Indeed, the hierarchical control structure relies on the optimization component $\boldsymbol{u_j}$ in order to avoid saturation at steady state. However, due to the dynamic nature of the optimization, saturation may still be reached during transients, as highlighted in remark \ref{remark1}, due to too fast trajectories or too aggressive control parameters tuning, potentially leading to error divergence. To address this issue, it would be necessary to break the independence between the high-level controller and the allocator, to allow the latter to change the control parameters in order to adjust the convergence rate so as to avoid saturation during transients, in a similar fashion to the well-known anti-windup mechanism.

\end{document}